\documentclass[conference]{IEEEtran} 

\usepackage{cite}
\usepackage{amsmath,amssymb,amsfonts}
\usepackage{algorithmic}
\usepackage{graphicx}
\usepackage{textcomp}
\usepackage{xcolor}
\usepackage[caption=false,font=footnotesize]{subfig}
\def\BibTeX{{\rm B\kern-.05em{\sc i\kern-.025em b}\kern-.08em
    T\kern-.1667em\lower.7ex\hbox{E}\kern-.125emX}}
\usepackage{url}
\usepackage{hyperref}
\usepackage{multirow}
\begin{document}

\title{ 
Illumination-Aware Contactless Fingerprint Spoof Detection via Paired Flash–Non-Flash Imaging
}


\author{
\IEEEauthorblockN{1\textsuperscript{st} Roja Sahoo}
\IEEEauthorblockA{
\textit{IIIT Hyderabad} \\
Gachibowli, India \\
roja.sahoo@research.iiit.ac.in
}
\and
\IEEEauthorblockN{2\textsuperscript{nd} Anoop Namboodiri}
\IEEEauthorblockA{
\textit{IIIT Hyderabad} \\
Gachibowli, India \\
anoop@iiit.ac.in
}
}


\maketitle

\begin{abstract}
Contactless fingerprint recognition enables hygienic and convenient biometric authentication but poses new challenges for spoof detection due to the absence of physical contact and traditional liveness cues. Most existing methods rely on single-image acquisition and appearance-based features, which often generalize poorly across devices, capture conditions, and spoof materials. In this work, we study paired flash–non-flash contactless fingerprint acquisition as a lightweight active sensing mechanism for spoof detection. Through a preliminary empirical analysis, we show that flash illumination accentuates material- and structure-dependent properties, including ridge visibility, subsurface scattering, micro-geometry, and surface oils, while non-flash images provide a baseline appearance context. We analyze lighting-induced differences using interpretable metrics such as inter-channel correlation, specular reflection characteristics, texture realism, and differential imaging. These complementary features help discriminate genuine fingerprints from printed, digital, and molded presentation attacks. We further examine the limitations of paired acquisition, including sensitivity to imaging settings, dataset scale, and emerging high-fidelity spoofs. Our findings demonstrate the potential of illumination-aware analysis to improve robustness and interpretability in contactless fingerprint presentation attack detection, motivating future work on paired acquisition and physics-informed feature design. Code is available in the repository.\footnotemark
\end{abstract}


\footnotetext{\url{https://github.com/clspooffnf/CLSpoofFNF}}

\section{Introduction and Motivation}
Contact-based fingerprint recognition is widely used but suffers from limitations such as hygiene concerns, sensor wear, and vulnerability to spoofing through physical replicas and latent prints. Contactless fingerprint acquisition addresses several of these issues by enabling touch-free capture and improved usability. However, the lack of physical contact introduces new challenges, particularly in reliably extracting 3D ridge and minutiae information under unconstrained 2D imaging conditions. It also removes many implicit liveness cues commonly exploited in contact-based systems–such as perspiration patterns, skin deformation, and electrical or capacitive conductivity used in optical and capacitive sensors–making it difficult to distinguish genuine fingerprints from visually similar spoof artifacts.

Most existing contactless fingerprint spoof detection approaches rely on a single captured image with flash. As a result they often depend on visual markers that demonstrate reduced robustness across changes in acquisition methods and spoof materials. In contrast, we consider paired flash–non-flash capture using a contactless smartphone camera, which provides controlled variation in illumination while keeping the fingerprint identity fixed. While lighting properties have been explored in related biometric contexts~\cite{9419913}, its use in contactless fingerprint spoof detection remains largely underexplored.

The goal of this work is to perform a preliminary analysis of the complementary information revealed by flash and non-flash captures, to understand the distinct features captured by each modality, and to explore how paired illumination can be leveraged for spoof detection. This study enables a single acquisition to provide dual information, combining passive information under ambient lighting with active response to controlled flash lighting.

The central question we explore is: \textit{Can illumination-induced variations in contactless fingerprint images be exploited to aid spoof detection?}

We argue that flash–non-flash capture should be viewed as a lightweight active sensing mechanism that exposes substrate-dependent illumination responses, enabling spoof detection beyond static appearance-based markers.

\section{Background and Related Work}

The development of contactless fingerprint recognition and presentation attack detection (PAD) has been supported by several publicly available datasets, though these remain limited in scale and diversity compared to contact-based benchmarks. Most existing contactless PAD datasets focus on print, replay, and material-based spoofing, typically captured using smartphone cameras under ambient illumination. Notable examples include \textit{COLFISPOOF}~\cite{10031150}, a large-scale benchmark with diverse print and replay presentation attack instruments (PAIs) and standardized evaluation protocols, and the \textit{IIITD Spoofed Fingerphoto Database}~\cite{7791201}, which provides early insights into print- and photo-based attacks. More recent datasets such as \textit{CLARKSON}~\cite{purnapatra2023presentationattackdetectionadvanced} expand spoof diversity by incorporating multiple PAI materials, synthetic (deepfake) fingerprints, and evaluations on unseen spoof types.

Beyond spoof-focused datasets, several large-scale contactless fingerprint datasets support matching and interoperability research. From IIITD, the  \textit{Smartphone Finger-Selfie Database v2 (ISPFDv2)}~\cite{9107238} and  \textit{UNFIT}~\cite{8575242} provide extensive collections captured under varied backgrounds and lighting conditions.  \textit{RidgeBase}~\cite{Jawade_2022} and the  \textit{PolyU Contactless-to-Contact Database}~\cite{8244291} further enable cross-sensor and contact-to-contactless matching by offering paired contactless and contact-based fingerprints acquired using multiple devices and lighting conditions. Still, most datasets rely on \textit{single-modality acquisition} and remain limited in size. Recent work on paired flash–non-flash capture~\cite{sahoo2026fusion2print} highlights controlled lighting as an additional sensing dimension. Consequently, illumination-dependent properties and composition-specific responses are underexplored, and sophisticated attacks such as high-fidelity 3D replicas and advanced display-based spoofs are insufficiently represented.

Early smartphone-based PAD methods used \textit{challenge-response} schemes based on specular reflections to separate real fingers from prints or simple fakes~\cite{6617150}. Hardware-assisted systems such as RaspiReader~\cite{engelsma2017raspireaderopensourcefingerprint} combine direct imaging with frustrated total internal
reflection (FTIR) but remain vulnerable to high-quality prosthetic replicas that closely mimic skin optics. Deep learning approaches, including semi-supervised ResNet-18 trained on synthetic fingerprints~\cite{adami2023universalantispoofingapproachcontactless} and GRU-AUNet with domain adaptation and attention~\cite{adami2025gruaunetdomainadaptationframework}, improve cross-spoof generalization. Feature-based methods like second-order local structures, color-space fusion, and CNN-based minutiae patch classification are lightweight but sensitive to material, camera, and lighting variations~\cite{8706242, app122211409, CNNmitigation2020}. Unsupervised autoencoder-based methods with attention further show promise for detecting unseen spoof types~\cite{adami2023contactlessfingerprintbiometricantispoofing}.

While most PAD methods rely on appearance-based markers, illumination-sensitive features have shown potential for contactless fingerprint spoof detection. CLNet~\cite{rajaram2024clnet} uses transfer learning to classify contactless fingerprints across multiple spoof types. Multispectral and polarization imaging exploit material responses under different lighting~\cite{abramovich2010spoof}, while texture-based features such as Local Binary Patterns (LBP) and wavelet energy capture structural differences~\cite{4579985}. These cues suggest that controlled lighting variations can reveal material-dependent differences that are difficult for spoof artifacts to replicate, motivating further investigation into paired flash–non-flash imaging for spoof detection.

\section{Empirical Observations Under Controlled Illumination}
This section analyzes the impact of paired flash–non-flash illumination on contactless fingerprint images using quantitative image quality metrics and model interpretability. 

We evaluated our approaches on a private dataset (due to biometric privacy constraints) containing 800 genuine fingerprint images and 1600 spoof samples–including digital replay (HD laptop screen) and colour print attacks–collected from 20 diverse individuals across two sessions. Images were captured using a Samsung Galaxy A54 smartphone with a custom app that records consecutive flash and non-flash image pairs, following the FNF Database~\cite{sahoo2026fusion2print} protocol.


\subsection{Photometric and Quality Variations Across Lighting Conditions}
The paired flash–non-flash contactless fingerprint captures introduce measurable differences in both image quality and photometric characteristics, which can provide valuable indicators for spoof detection. To quantify these variations, we evaluate several established and custom image quality metrics that describe ridge clarity, local texture, and RGB color distribution statistics over the images.

\begin{figure}[htbp]
    \centering
    \includegraphics[width=0.84\linewidth]{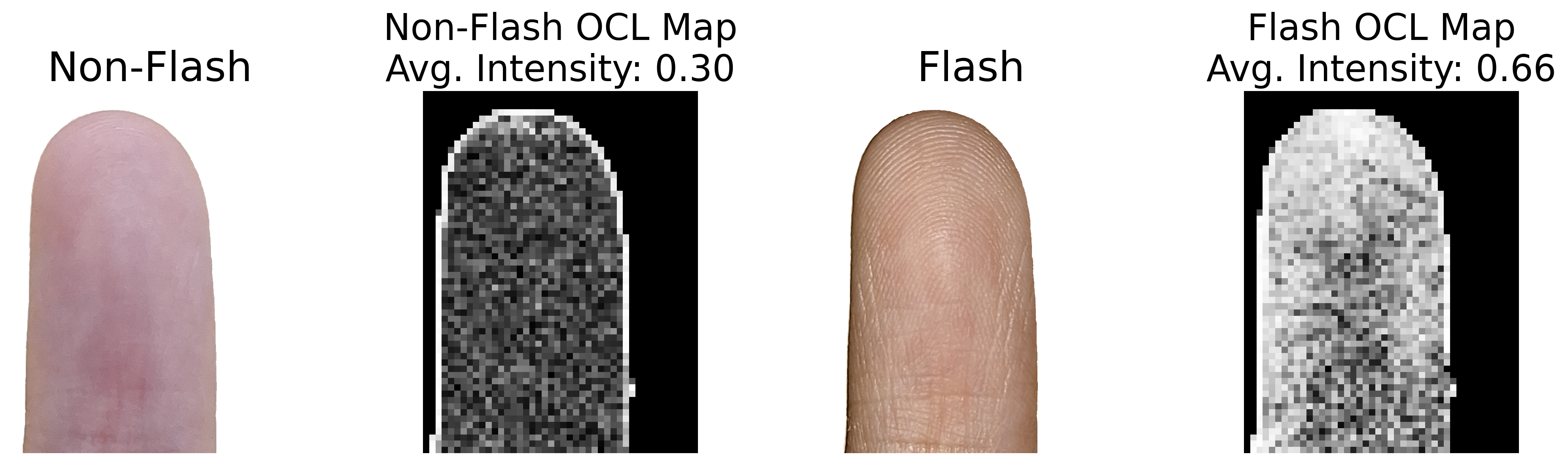}
    \caption{Blockwise OCL maps for sample flash and non-flash fingerprint images. Brighter regions(higher intensity) indicate higher ridge clarity; flash images show consistently stronger ridge coherence.}
    \label{fig:1}
\end{figure}

\begin{figure}[htbp]
    \centering
    \includegraphics[width=0.84\linewidth]{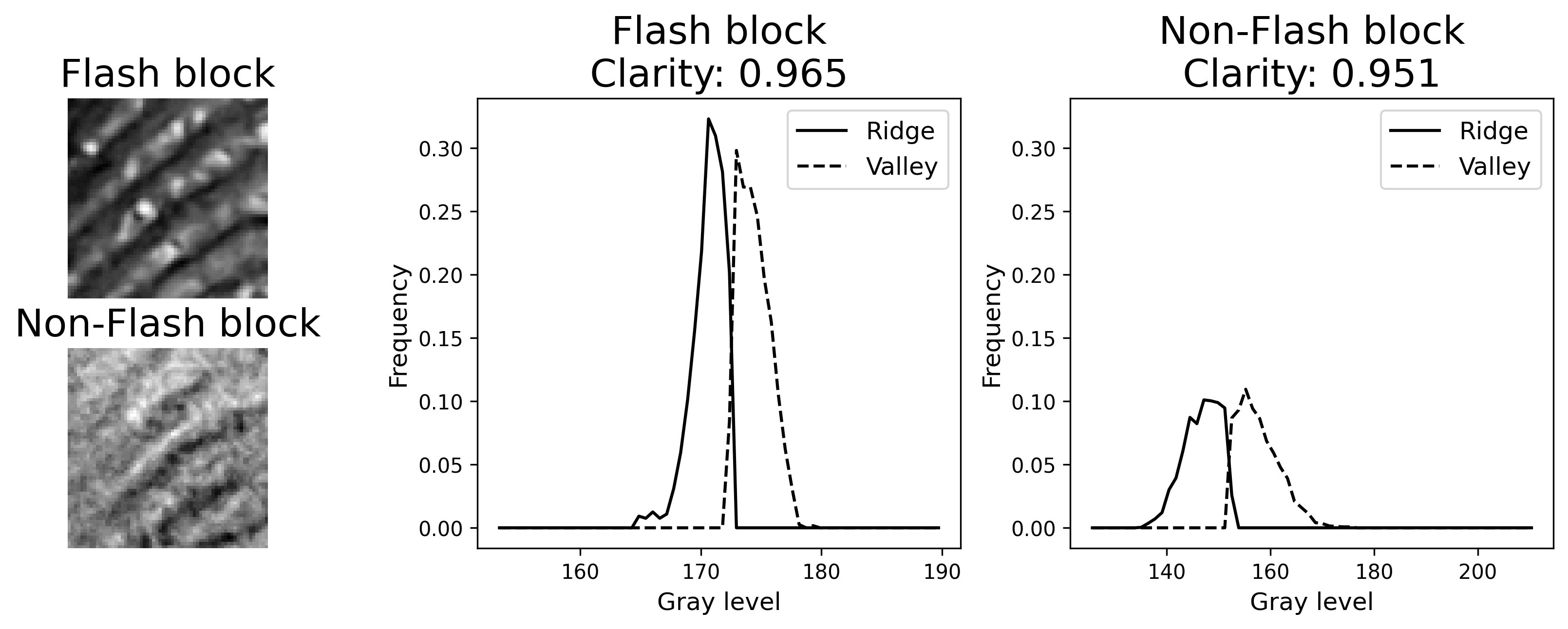}
    \caption{LCS and ridge-valley intensity profiles for sample flash and non-flash fingerprint blocks, showing improved ridge-valley contrast under flash illumination.}
    \label{fig:2}
\end{figure}

\begin{figure}[htbp]
    \centering
    \includegraphics[width=0.84\linewidth]{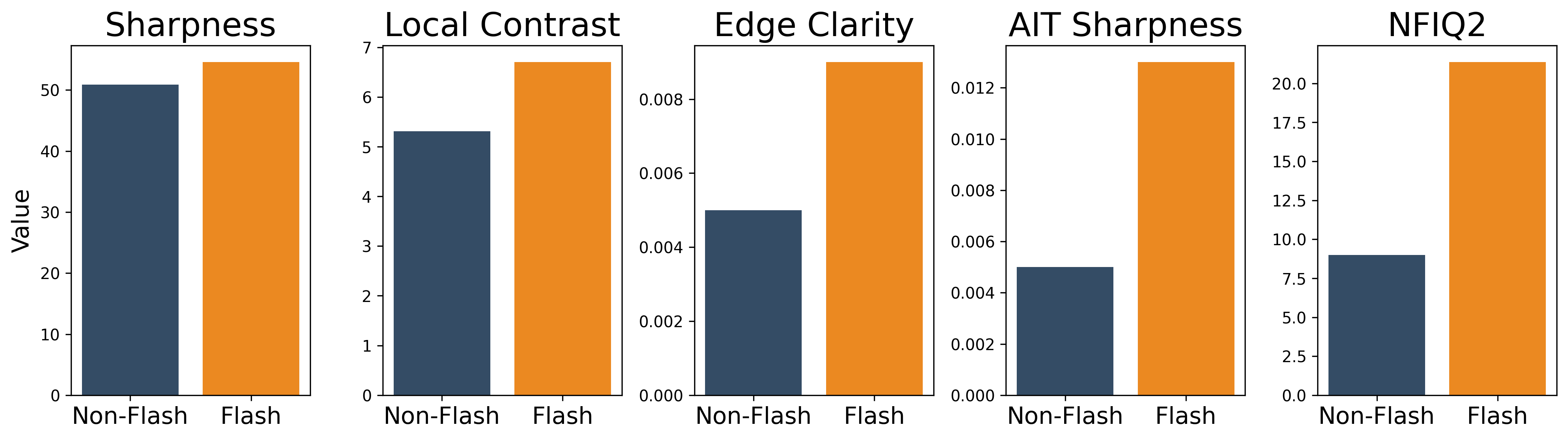}
    \caption{Comparison of standard (AIT Sharpness, NFIQ2) and custom patch-wise image quality metrics over the sample dataset, where flash images outperform non-flash images across all measures.}
    \label{fig:3}
\end{figure}

For overall ridge clarity, we compute the \textit{Orientation Certainty Level (OCL)}~\cite{1038062}, which measures the local directional consistency of ridge structures, and the \textit{Local Clarity Score (LCS)}~\cite{Alonso_Fernandez_2007}, which captures the perceptual clarity of ridge-valley regions. Additionally, the \textit{NIST Fingerprint Image Quality 2 (NFIQ2)}~\cite{tabassi2021nistir8382} metric provides a standardized assessment of image quality from multiple aspects of ridge definition and minutiae reliability. We also measure AIT sharpness~\cite{s21072248}, along with locally computed contrast, edge clarity and sharpness metrics derived from our own implementations. These metrics collectively provide a quantitative characterization of how flash illumination affects ridge visibility, texture prominence, and overall fingerprint quality compared to non-flash captures shown in Figures~\ref{fig:1}–\ref{fig:3}.

\begin{figure}[htbp]
    \centering
    \includegraphics[width=0.84\linewidth]{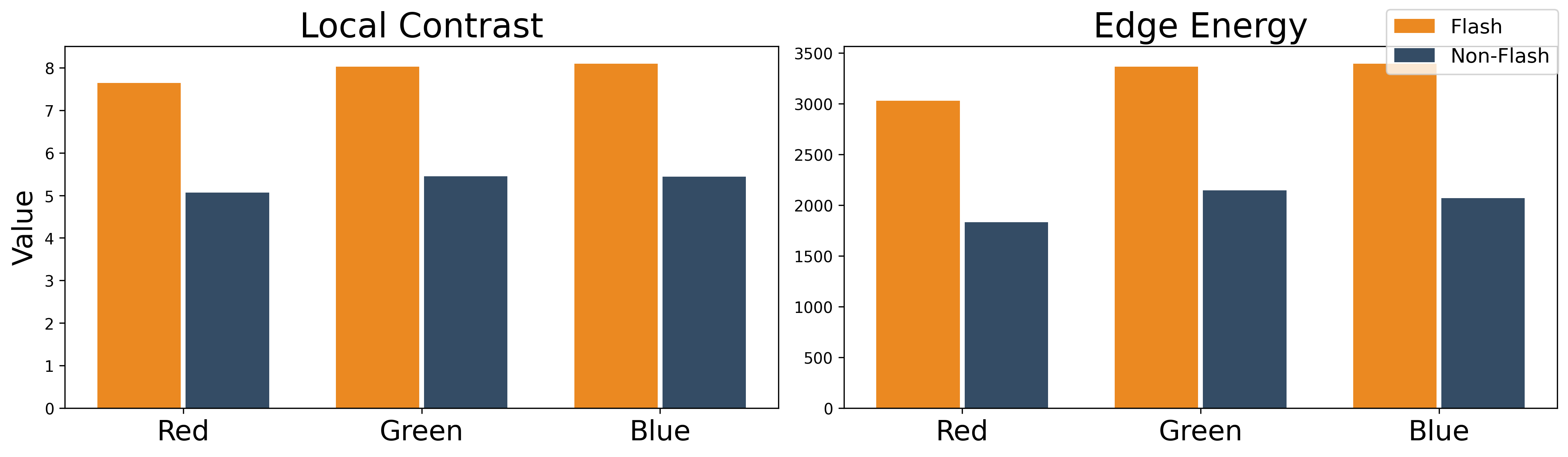}
    \caption{Photometric analysis of local contrast and edge energy for R, G, and B channels of the sample dataset under flash and non-flash illumination.}
    \label{fig:4}
\end{figure}

In addition to image quality, we analyze photometric variations at the channel level in Figure~\ref{fig:4}. Metrics include the \textit{local contrast} (average standard deviation of pixel intensities of non-overlapping patches) and the \textit{edge energy} (mean squared Sobel gradient magnitude) of each RGB channel, which together reveal illumination-dependent variations in material reflectance and ridge definition. Flash lighting often introduces stronger highlights and contrast, particularly along ridges, whereas non-flash captures provide a more uniform appearance across channels. The differences between channels and across illumination conditions serve as an informative vector for differentiating genuine fingerprints from spoof artifacts.

Collectively, these image quality and photometric measures provide a detailed view of how controlled illumination variations affect fingerprint appearance, offering both baseline quality assessment and material-dependent cues that can aid in spoof detection.

\subsection{Model Interpretability Under Illumination Change}

Understanding how deep models focus on fingerprint structures under varying lighting can provide valuable insights for spoof detection. We analyze spatial attention maps across flash and non-flash images using two architectures, including a general transformer DINOv2~\cite{oquab2024dinov2learningrobustvisual} (not fine-tuned on fingerprints) and a ResNet-18~\cite{he2015deepresiduallearningimage} with an embedding head, both with and without fingerprint-specific fine-tuning. 

\begin{figure}[htbp]
    \centering

    \subfloat[Non-Flash with DINOv2]{%
        \includegraphics[width=0.66\linewidth]{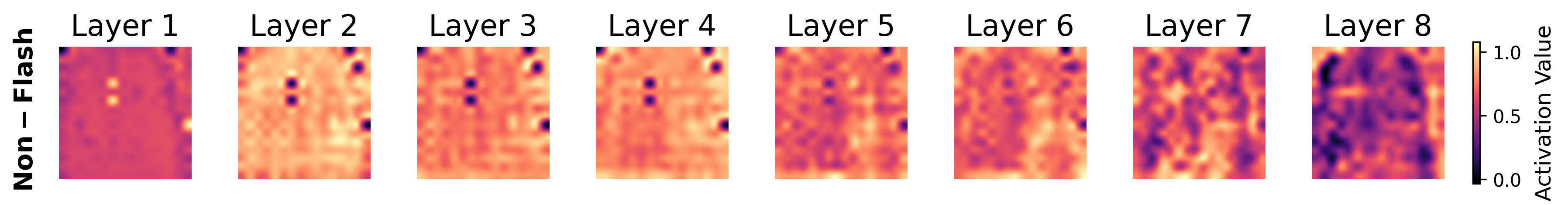}
        \label{fig:5a}
    }
    \vspace{0.1em}
    \subfloat[Flash with DINOv2]{%
        \includegraphics[width=0.66\linewidth]{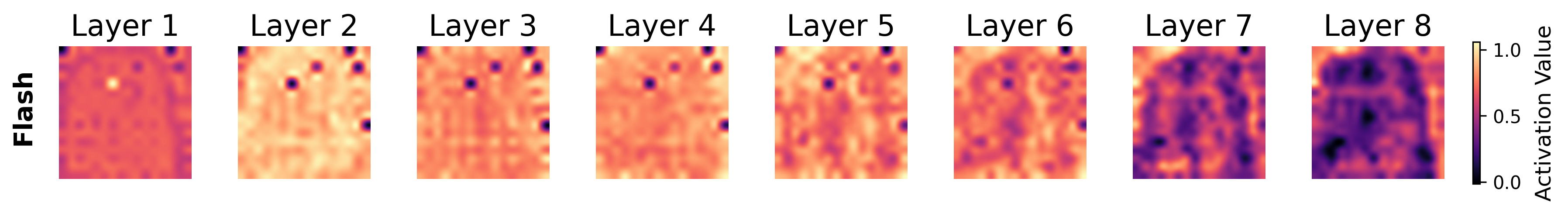}
        \label{fig:5b}
    }


    \subfloat[Non-flash with ResNet-18 based model]{%
        \includegraphics[width=0.66\linewidth]{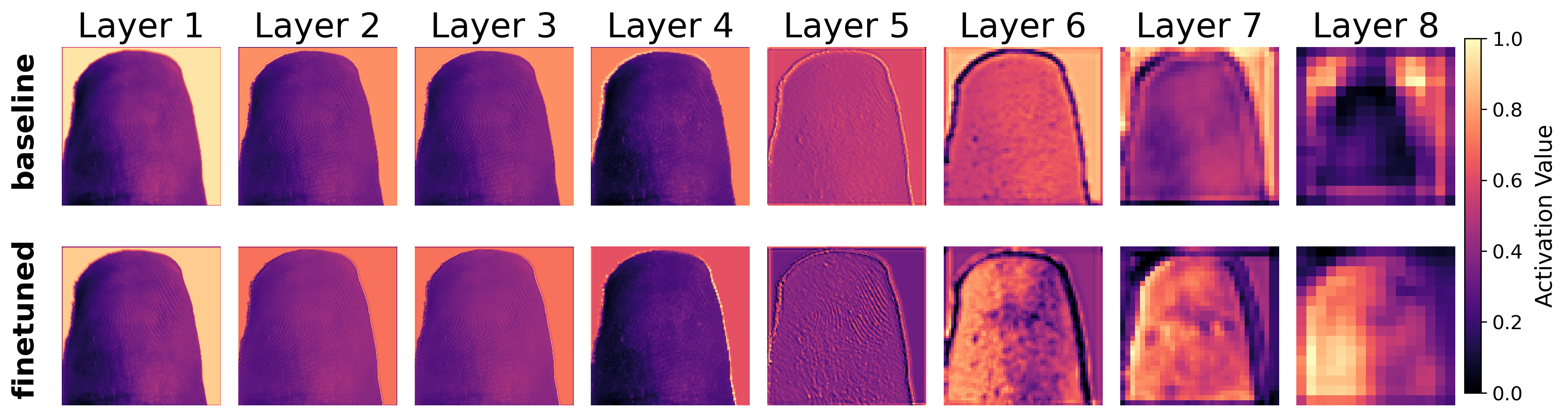}
        \label{fig:5c}
    }
    \vspace{0.1em}
    \subfloat[Flash with ResNet-18 based model]{%
        \includegraphics[width=0.66\linewidth]{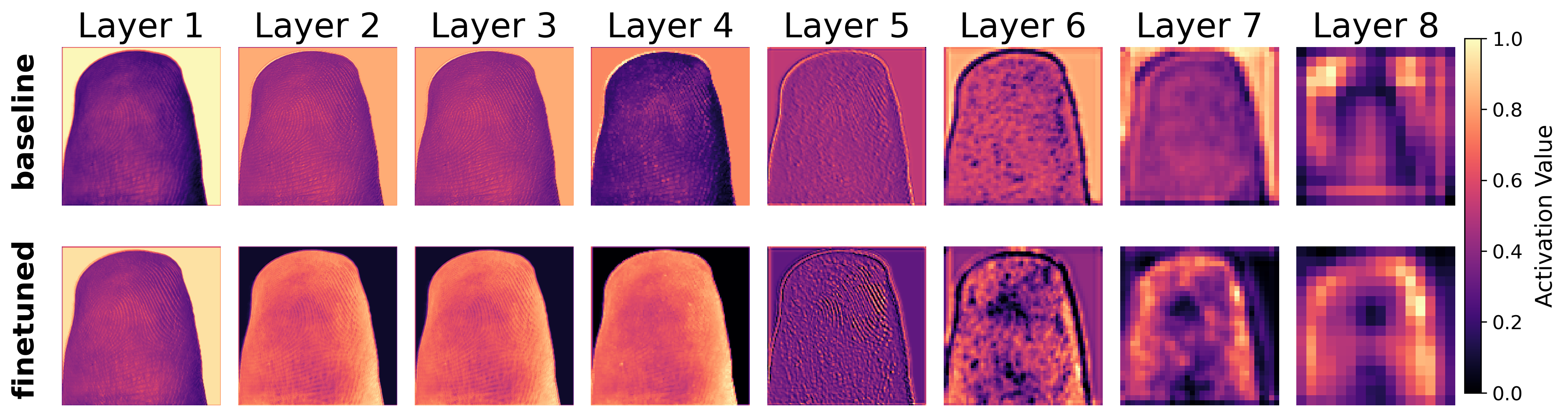}
        \label{fig:5d}
    }

    \caption{Attention maps for sample flash and non-flash contactless fingerprint images. (a)–(b) Baseline DINOv2 attention maps with limited ridge awareness but improved spatial structure under flash lighting. (c)–(d) Fine-tuned ResNet-18–based model showing enhanced ridge-focused attention, especially for flash images.}

    \label{fig:5}
\end{figure}

Without fine-tuning, models primarily capture coarse fingerprint regions without detailed ridge-level understanding. Attention is broadly distributed across the fingerprint area and is relatively similar for both flash and non-flash images. When fine-tuned on fingerprints the models begin to highlight ridge structures more consistently, illustrated in Figure~\ref{fig:5}. Notably, flash images tend to elicit stronger attention on these critical ridge regions, suggesting that controlled lighting enhances ridge visibility and the model's focus on informative micro-structures. Non-flash images, while still highlighting fingerprint regions, generally produce weaker and more diffuse attention maps.

Further quantitative analysis in Table~\ref{tab:1} demonstrates improved genuine–spoof class separation, with Fisher Discriminant Ratios increasing from $5\mathrm{e}{-4}$ to $0.18$ for non-flash and $7\mathrm{e}{-4}$ to $1.48$  for flash images, indicating stronger discrimination after fine-tuning.

\begin{table}[htbp]
\centering
\caption{Activation statistics for genuine (G) and spoof (S) fingerprints under flash and non-flash conditions for the ResNet-18–based model, computed over the sample dataset (mean $\pm$ std).}
\label{tab:1}
\setlength{\tabcolsep}{4pt}
\renewcommand{\arraystretch}{1.15}
\begin{tabular}{l c c c c}
\hline
\textbf{Model} & \textbf{G-Non-Flash} & \textbf{G-Flash} & \textbf{S-Non-Flash} & \textbf{S-Flash} \\
\hline
Baseline   
& $0.21 \pm 0.19$ 
& $\mathbf{0.23 \pm 0.16}$ 
& $0.22 \pm 0.24$ 
& $0.23 \pm 0.21$ \\

Fine-tuned 
& $0.37 \pm 0.16$ 
& $\mathbf{0.49 \pm 0.13}$ 
& $0.26 \pm 0.22$ 
& $0.27 \pm 0.14$ \\
\hline
\end{tabular}
\end{table}

\section{Discussion}
We discuss potential adversarial presentation attacks in contactless fingerprint recognition and examine how spoof detection systems can be designed using flash–non-flash lighting principles to counter them. We also outline the limitations of illumination-based cues and highlight practical challenges that may arise in real-world deployment.

\begin{table*}[htbp]
\centering
\caption{Summary of contactless fingerprint PAIs and illumination-dependent cues exposed by paired flash–non-flash acquisition.}
\label{tab:2}
\renewcommand{\arraystretch}{1.15}
\begin{tabular}{p{2.5cm} p{3.2cm} p{4.2cm} p{4.2cm}}
\hline
\textbf{PAI Type} & \textbf{Typical Forms} & \textbf{Material / Method Weakness} & \textbf{Cues Exposed by Flash–Non-Flash} \\
\hline

Printed spoofs &
Paper printouts (inkjet, laser) &
Flat surface, no subsurface scattering; ink-dependent color response; uniform reflectance &
Exaggerated specular highlights under flash; ink-induced RGB distortions; low micro-texture consistency across illumination \\

Digital spoofs &
Smartphone / tablet displays &
Emissive surface; pixel grid structure; channel saturation &
Screen glare and moir\'e patterns under flash; unnatural channel peaks; inconsistent photometric response \\

Physical / molded spoofs &
Silicone, latex, gelatin, resin replicas; prosthetic fingers (closely matching human skin tone) &
Surface-only reflection; limited or no subsurface scattering; uniform ridge depth &
Uniform specular reflection under flash; reduced ridge-valley micro-variation; inconsistent differential response \\

3D modeled spoofs &
3D replicas generated from 2D contactless images &
Synthetic geometry; missing organic ridge irregularities; material homogeneity &
Illumination-invariant texture patterns; weak flash–non-flash differences; absence of oil- or sweat-induced highlights \\

\hline
\end{tabular}
\end{table*}

\subsection{Presentation Attack Scenarios}
In contactless fingerprint recognition, the absence of physical contact allows a wide range of PAIs, each interacting differently with illumination. Broadly, these attacks can be categorized into printed spoofs, digital display-based spoofs, and physical or molded replicas which have also been summarized in Table~\ref{tab:2}.

\textbf{Printed spoofs}, such as inkjet or laser printouts on paper, can resemble genuine fingerprints under non-flash illumination, particularly in ridge layout and global appearance. Flash lighting, however, exposes flat surface reflectance, ink-dependent color distortions, exaggerated specular highlights, and printing artifacts. The lack of subsurface scattering and uniform surface response of paper-based materials results in illumination behavior that differs significantly from real skin when comparing flash and non-flash captures.

\textbf{Digital spoofs} presented on smartphones or tablets may appear realistic under ambient lighting due to high-resolution displays and controlled rendering. Under flash lighting, characteristic artifacts such as screen glare, pixel grid or moiré patterns~\cite{niu2021moireattackmanew}, and unnatural RGB channel saturation become apparent. These displays exhibit illumination responses and channel-specific peaks that are inconsistent with the photometric behavior of human skin.

\textbf{Physical or molded spoofs} include three-dimensional replicas made from silicone, latex, gelatin, resin, or prosthetic fingers, often designed to closely match human skin tone. These replicas may reproduce ridge depth and shading under non-flash illumination and can be fabricated either from physical molds or increasingly from 3D models derived from 2D contactless images~\cite{8100983}. Flash lighting highlights key composition differences, including uniform specular reflection, reduced micro-texture variation, and the absence of true subsurface scattering and natural oil distribution, leading to inconsistent ridge-valley transitions across illumination conditions.

Across all PAI types, paired flash–non-flash acquisition introduces a controlled lighting change that exposes identifiable weaknesses. While spoof artifacts may closely match genuine fingerprints under a single illumination condition, their differential response to flash provides discriminative features that can be leveraged for robust spoof detection.

\subsection{Implications for Spoof Detection Design}

Contactless fingerprint spoof detection can leverage the complementary information revealed by paired flash–non-flash acquisition. By analyzing fingerprint responses under controlled illumination changes, it is possible to extract \textit{material- and structure-dependent features} that are difficult for spoof artifacts to replicate. 
Using both flash and non-flash images enables models to capture substrate-specific behaviors, as exposure-dependent changes in reflection and texture reveal rich markers for spoof detection.
In the following subsections, we discuss key conceptual cues and their potential applications with results from our dataset.

\begin{figure}[htbp]
    \centering

    \subfloat[Pearson correlation heatmaps]{%
        \includegraphics[width=0.7\linewidth]{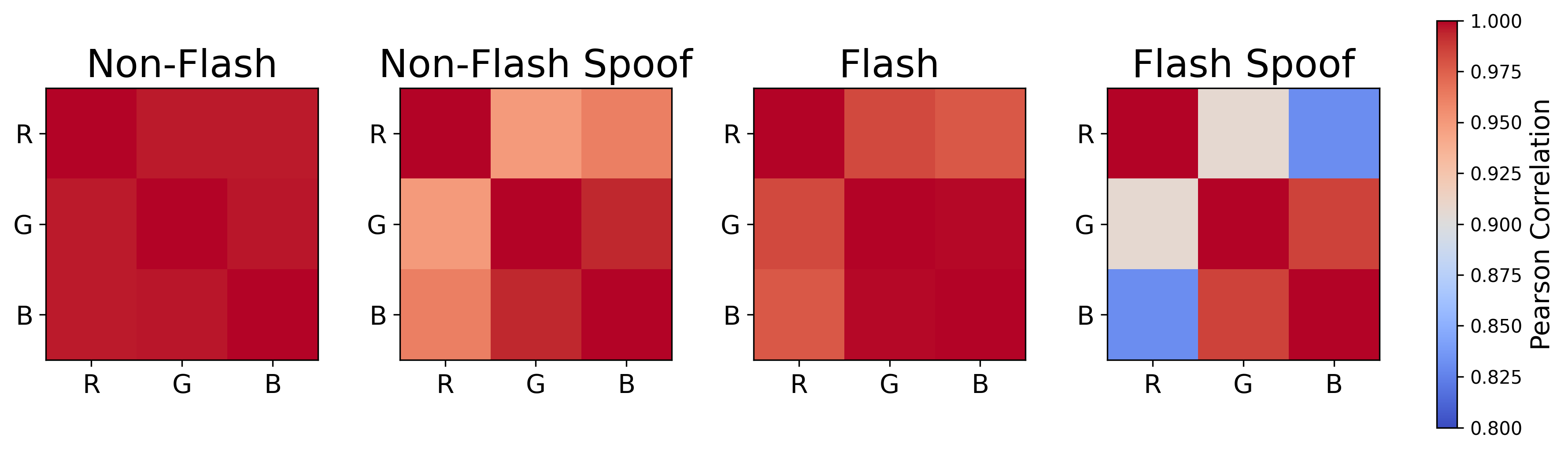}
        \label{fig:6a}
    }
    \hfill
    \subfloat[Mutual information heatmaps]{%
        \includegraphics[width=0.7\linewidth]{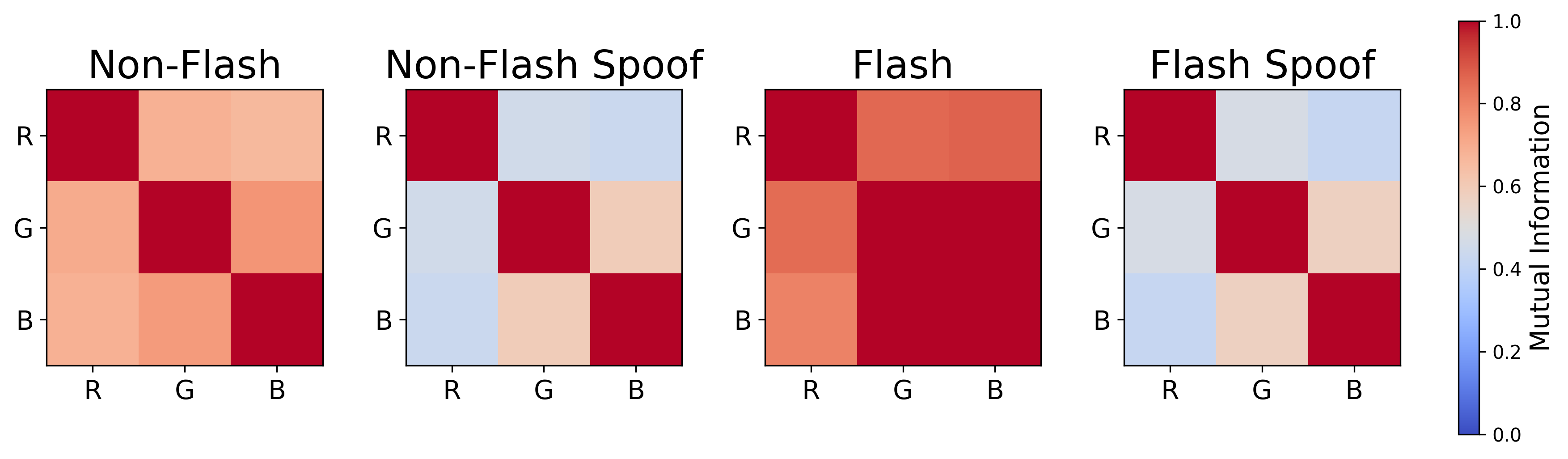}
        \label{fig:6b}
    }
    \caption{Inter-channel correlation analysis for the sample dataset of flash and non-flash fingerprint images and their spoof counterparts. (a) Pearson correlation and (b) mutual information heatmaps show stronger off-diagonal channel relationships for genuine images, with clearer distinction from spoof under flash illumination. 
    Mann–Whitney U tests confirm the statistical significance of off-diagonal activations (Pearson: $p < 0.001$ for both conditions; MI: $p < 0.001$ for flash and $p = 0.035$ for non-flash).}
    \label{fig:6}
\end{figure}

\textbf{1. Inter-channel Correlation:} One informative cue for distinguishing genuine fingerprints from spoofs is the statistical relationship between the RGB channels (typical for human skin colours), which can be measured using Pearson correlation or mutual information (MI), e.g., $\text{corr}(R,G)$, $\text{corr}(R,B)$, and $\text{corr}(G,B)$. Real skin exhibits high inter-channel correlation due to uniform subsurface scattering and pigmentation, whereas spoofs introduce channel decorrelation as shown in Figure~\ref{fig:6}. Screens-based spoofs often show unnatural peaks in specific channels and prints display inconsistent RGB ratios due to ink properties. Flash illumination amplifies composition-related reflection differences, making decorrelations more pronounced and easier to detect (with Pearson: $\Delta_\text{flash} = 0.090$, $\Delta_\text{non-flash} = 0.032$; MI: $\Delta_\text{flash} = 0.167$, $\Delta_\text{non-flash} = 0.041$, where $\Delta$ denotes the genuine–spoof separation in inter-channel correlation statistics).

\textbf{2. Specular Reflection Features:} Specular reflection patterns under flash illumination provide another rich set of cues. Metrics include the specular highlight ratio (percentage of pixels with high intensity across RGB channels but low local texture), channel saturation under flash (clipped channels), and color temperature shifts, which can be captured as changes in R/B ratios. Genuine skin exhibits controlled specular reflections along ridges due to ridge curvature and surface oils, whereas printed spoofs produce exaggerated highlights and screen-based spoofs exhibit unnatural RGB peaks, as computed in Table~\ref{tab:3}. Comparing flash and non-flash responses enhances these differences, as flash selectively emphasizes material-dependent reflectivity.

\begin{table}[htbp]
\centering
\caption{Average Specular Highlight Ratio (SHR) over the dataset of real and spoof fingerprints using paired flash and non-flash images. Higher SHR indicates increased specular reflection and is more indicative of spoof fingerprints.}
\label{tab:3}
\renewcommand{\arraystretch}{1.15}
\begin{tabular}{l c}
\hline
\textbf{Sample Type} & \textbf{SHR} \\
\hline
Real Fingerprints  & 0.009 \\
Spoof Fingerprints & 0.043 \\
\hline
\end{tabular}
\end{table}

\begin{figure}[htbp]
    \centering

    \subfloat[Texture Realism analysis for Flash]{%
        \includegraphics[width=0.7\linewidth]{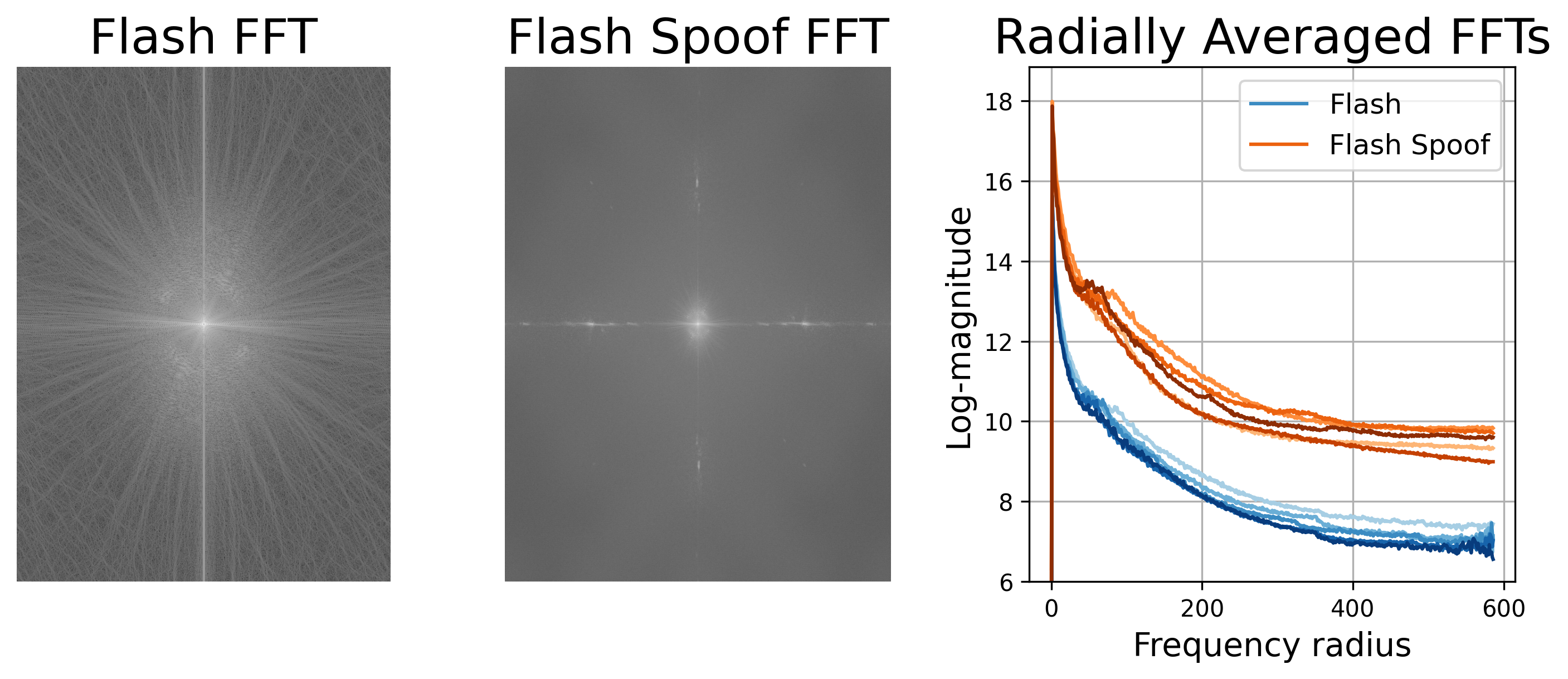}
        \label{fig:7a}
    }
    \hfill
    \subfloat[Texture Realism analysis for Non-Flash]{%
        \includegraphics[width=0.7\linewidth]{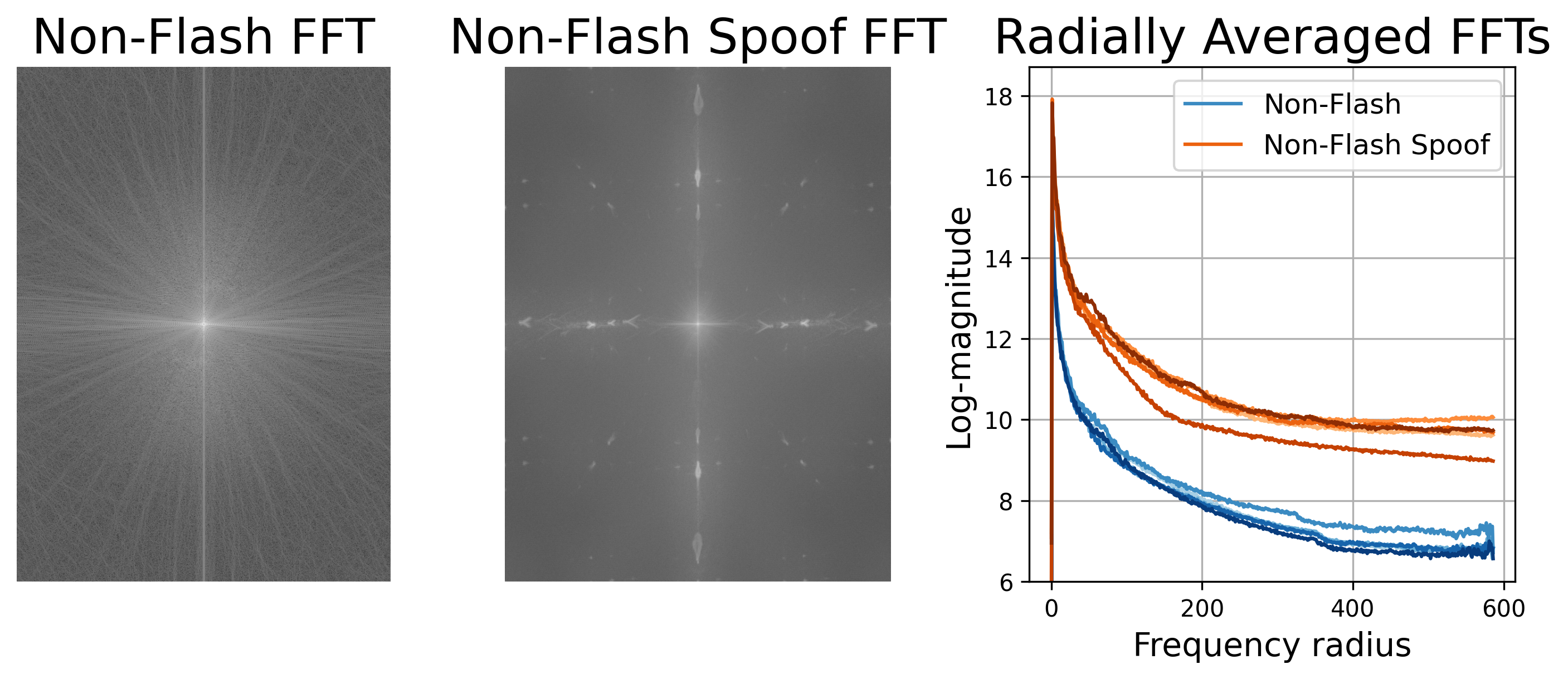}
        \label{fig:7b}
    }
    \caption{Log-magnitude FFTs and radially averaged spectra are shown for (a) flash and (b) non-flash contactless fingerprint images of genuine and spoof samples. Texture realism ratio (fraction of blocks with dominant FFT peaks) of the dataset is higher for spoof fingerprints (0.226) than real fingerprints (0.087), indicating disrupted, grid-like textures and increased high-frequency artifacts enabling a clear distinction.}
    \label{fig:7}
\end{figure}

\textbf{3. Texture Distribution and Realism:} Texture analysis provides another axis for differentiating genuine fingerprints from spoofs. Descriptors such as LBP~\cite{sedaghatjoo2024localbinarypatternlbpoptimization}, Gray Level Co-occurrence Matrices (GLCM)~\cite{zubair2024greylevelcooccurrencematrix}, and Fourier spectrum analysis reveal unnatural ridge-valley periodicity and repetitive patterns commonly found in spoof materials. Real fingerprint ridges exhibit organic variability, whereas spoofs tend to have smoother, repetitive structures due to casting or printing processes, as shown in Figure~\ref{fig:7}. By analyzing differences in texture descriptors across flash and non-flash images (e.g., $\Delta$LBP, $\Delta$GLCM, or $\Delta$Fourier transforms), illumination-dependent texture stability can be quantified as an indicator of substrate authenticity.

\textbf{4. Differential Imaging (Flash minus Non-Flash):} Rather than relying solely on absolute appearance, analyzing the difference between flash and non-flash images isolates material- and geometry-dependent responses. Differential imaging suppresses ambient lighting while emphasizing required information. Spoofs typically show weak or inconsistent changes across illumination changes due to the lack of subsurface scattering or surface heterogeneity. This flash minus non-flash construct enables learning \textit{physics-informed features} beyond plain appearance-based indicators.

\textbf{5. Subsurface Scattering:} Human skin exhibits subsurface scattering (SSS), where light penetrates and re-emerges from beneath the surface~\cite{ANDERSON198113}. Spoof materials are predominantly surface-reflective and lack this effect. Under flash illumination, SSS produces smoother ridge-valley transitions in real fingers, while spoofs exhibit sharper or more uniform transitions. This can be quantified using local intensity distributions or gradient smoothness across flash and non-flash captures.

\textbf{6. Micro-Geometry:} Micro-geometry captures fine ridge-valley structure, including ridge depth variability and slope irregularities. Genuine fingerprints show organic, non-uniform slopes and varying ridge heights, whereas spoof artifacts exhibit uniform, repetitive structures from casting or printing. Shadow patterns and differential imaging reveal these inconsistencies, which are difficult for spoofs to replicate.

\textbf{7. Surface Oils and Sweat:} Natural oils and sweat create irregular micro-reflective highlights irregularly distributed on genuine skin, particularly under flash illumination. Spoof artifacts generally lack such fine-grained features, producing no highlights or overly uniform reflections. Flash–non-flash comparison accentuates these differences by selectively highlighting real skin micro-reflectivity.

\subsection{Limitations and Challenges}

Despite the advantages of paired flash–non-flash imaging for contactless fingerprint spoof detection, several challenges remain. Variations in ambient lighting, finger pose, distance to the camera, finger size, and out-of-plane rotations can introduce inconsistencies, affecting feature stability. Smartphone captures are sensitive to sensor noise, motion blur, and temporal misalignment. Advanced spoofing techniques–such as high-quality 3D-printed replicas or screen-based spoofs with anti-reflection coatings–can partially mimic illumination-dependent responses, reducing the discriminative power of certain properties. Paired captures also require extra user cooperation and a more complex acquisition pipeline, impacting usability in large-scale or unconstrained deployments.

Learning robust models from paired exposure markers demands larger, more diverse datasets covering genuine fingerprints and emerging high-effort spoofs. Current datasets are limited in scale and diversity, constraining generalization. Moreover, most existing approaches rely on 2D imaging, missing 3D surface micro-geometry and leaving vulnerabilities against highly realistic replicas. These limitations highlight the need for expanded datasets, robust preprocessing, and models that generalize across devices, capture conditions, finger characteristics, and both 2D and 3D spoof types.

\section{Conclusion}

In this work, we presented an analysis of contactless fingerprint spoof detection using paired flash–non-flash captures, demonstrating that this setup provides complementary cues that are difficult for spoof artifacts to replicate. Flash lighting enhances ridge visibility and distinctive reflective properties, while non-flash images provide a reference visual appearance. Differential imaging and attention-based model interpretability further emphasize these differences, found to be statistically significant, enabling physics-informed feature extraction beyond traditional appearance-based methods.

We outlined design principles for spoof detection leveraging multiple indicative physical properties and illumination-induced features. Paired exposure improves interpretability and provides robust quantitative and qualitative indicators for detecting diverse spoof types, including printed, digital, and molded artifacts. Despite limitations in capture variability, dataset scale, and evolving spoof technologies, our findings demonstrate the potential of illumination-aware approaches to enhance the reliability and generalization of contactless fingerprint PAD systems.

\subsection*{Future Work}
Future work can focus on integrating these insights into end-to-end deep learning frameworks, expanding paired flash–non-flash datasets with spoof samples, and exploring real-time implementations on mobile and embedded devices.

\bibliographystyle{IEEEtran}
\bibliography{main}

\end{document}